\documentclass{article}
\usepackage{spconf,amsmath,graphicx}
\usepackage{bm}
\usepackage{url} 
\usepackage[ruled,vlined]{algorithm2e}
\usepackage{multirow}

\title{A HIERARCHICAL DECODING MODEL FOR SPOKEN LANGUAGE UNDERSTANDING FROM UNALIGNED DATA} 
\name{Zijian Zhao, Su Zhu and Kai Yu\thanks{The corresponding author is Kai Yu. This work has been supported by the China NSFC projects (No. 61573241). Experiments have been carried out on the PI supercomputer at Shanghai Jiao Tong University.}}

\address{MoE Key Lab of Artificial Intelligence\\
	SpeechLab, Department of Computer Science and Engineering\\
	Shanghai Jiao Tong University, Shanghai, China\\
    \texttt{\{1248uu, paul2204, kai.yu\}@sjtu.edu.cn}}

\begin{document}
%
\maketitle
\begin{abstract}
Spoken language understanding (SLU) systems can be trained on two types of labelled data: aligned or unaligned. Unaligned data do not require word by word annotation and is easier to be obtained. In the paper, we focus on spoken language understanding from unaligned data whose annotation is a set of act-slot-value triples. Previous works usually focus on improve slot-value pair prediction and estimate dialogue act types separately, which ignores the hierarchical structure of the act-slot-value triples. Here, we propose a novel hierarchical decoding model which dynamically parses act, slot and value in a structured way and employs pointer network to handle out-of-vocabulary (OOV) values. Experiments on DSTC2 dataset, a benchmark unaligned dataset, show that the proposed model not only outperforms previous state-of-the-art model, but also can be generalized effectively and efficiently to unseen act-slot type pairs and OOV values.  
\end{abstract}

\begin{keywords}
Spoken language understanding, unaligned data, hierarchical decoding, pointer network 
\end{keywords}
\section{INTRODUCTION}
\label{sec:intro}
The spoken language understanding (SLU) module is a key component of spoken dialogue system (SDS), parsing user's utterances into corresponding semantic forms. Typically, the SLU problem is regarded as a sequence tagging task which needs word-level annotations\cite{mesnil2013investigation,yao2013recurrent,zhu2016encoder}, e.g., the utterance \textit{``Show me flights from Boston to New York''} can be parsed as \textit{``Show me flights from [Boston:from\_city] to [New York:to\_city]''} \cite{pieraccini1992speech}. Beyond this word aligned annotation, there is also sentence-level semantic annotation which is unaligned, e.g., the utterance \textit{``I want a high priced restaurant''} has an act-slot-value triple annotation of \textit{``inform(pricerange=expensive)''} and the utterance \textit{``what type of food does it serves''} has an annotation of \textit{``request(food)''}. 

The unaligned SLU has some advantages against the aligned one. First, as a downstream module of Automatic Speech Recognition (ASR), SLU module based on statistical method often requires that training data should be labelled on the outputs from ASR, which can improve robustness to ASR errors. Therefore, it is hard and sometimes impossible to align the semantic annotations onto ASR outputs due to ASR errors (especially word insertion and deletion errors). Second, value aliases are also difficult to be handled in a word-aligned way which is very time-consuming. In this paper, we focus on SLU with the unaligned semantic annotation that a sentence is labelled as a set of act-slot-value triples \cite{young2007cued}.


There are numerous previous works for the unaligned SLU. Support vector machines (SVM) have been used for learning semantic tuple classifiers \cite{mairesse2009spoken,henderson2012discriminative}. Yazdani et al. propose a model to calculate the similarity between the input sentence and all possible semantic tuples \cite{yazdani2015model}. It assumes all possible values have been known, which may be impractical in real applications and inefficient, e.g. there are a large number of songs in a music domain. Sentence and context representations are exploited in \cite{barahona2016exploiting} and the OOV problem in slot values is addressed by utilizing a pointer network in \cite{zhao2018improving}. However, they predict act type and slot-value pair separately ignoring relation between the act and slot-value pair.


\begin{figure}[!htpb]
\centering
\includegraphics[width=1.0\linewidth,trim=.05cm .02cm .02cm .05cm, clip]{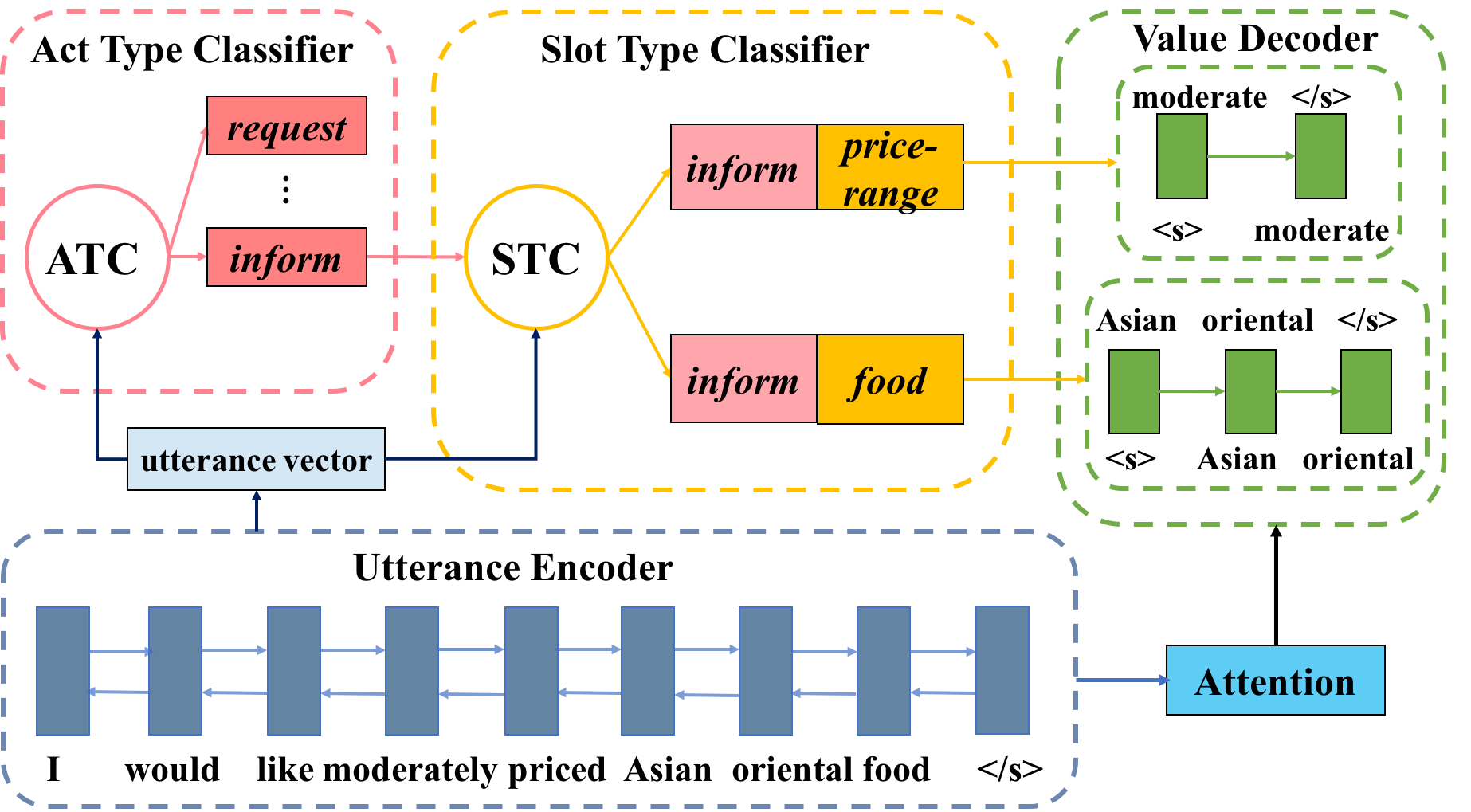}
\caption{A Hierarchical Decoding Model for Spoken Language Understanding from Unaligned Data. ATC denotes act type classifier and STC denotes slot type classifier.}
\label{fig:framework}\textsf{}
\end{figure}

In this paper, we propose a novel hierarchical decoding model for SLU from unaligned data. The model predicts act-slot-value triples hierarchically by following the triple structure. The hierarchical decoding can predict multiple act-slot-value triples completely and generalize to unseen act-slot type pairs. Pointer network \cite{vinyals2015pointer} is employed to generate out-of-vocabulary (OOV) values with a context-aware attention mechanism. In the experiments, our method achieves state-of-the-art performance in the DSTC2 dataset, and shows great generalization capacity.

The rest of the paper is organized as follows. The next section introduces relations to prior works. In section 3, we describe the hierarchical decoding model in detail. Experiments and analyses are presented in section 4, followed by conclusions.

\vspace{-2.5mm}
\section{RELATION TO PRIOR WORK}
The work most relevant to us is \cite{zhao2018improving}, which also uses a pointer network \cite{vinyals2015pointer} to handle the OOV problem of values. They focus on filling value for a slot which is a sub-task of SLU, while we aim to hierarchically generate a set of act-slot-value triples which is a complete target of the unaligned SLU. Moreover, the relation between act and slot-value pair is ignored in \cite{zhao2018improving}. We apply a context-aware attention mechanism within the pointer network by incorporating the predicted act and slot. A comparison of results on the DSTC2 dataset also shows that we can get better performance.

\section{HIERARCHICAL DECODING MODEL}

In this section, the details of our model are given. The task of SLU from unaligned data is to predict a set of act-slot-value triples given an input utterance. To represent rich semantics, the triples are of three types: single act like \textit{``thankyou()''} and \textit{``bye()''}, act-slot pair like \textit{``request(food)''}, and act-slot-value triple like \textit{``inform(pricerange=expensive)''}, which are given in domain ontology. Not all act types are followed by a slot and value. Thus, we predict the act-slot-value triples by following the triple structure. The overall model consists of four  modules, as shown in Figure \ref{fig:framework}:
\begin{itemize}
    \item a shared utterance encoder;
    \item an act type classifier with the utterance as input to predict act types;
    \item a slot type classifier with the utterance and an act type as inputs to predict slot types;
    \item a value decoder with the utterance and an act-slot type pair as inputs to generate the value sequence.
\end{itemize}

In training state, the modules can be trained together at the same time as multi-task learning. However, in testing stage, the modules decode recursively to generate the triples. The details of hierarchical decoding are given in Algorithm \ref{alg:decoding}.

\begin{algorithm}[h]
\SetAlgoLined
\KwIn{utterance, ontology, model}
\KwOut{a set of act-slot-value triples}
Initialize a empty list\;
pred\_acts = act\_classifier(utterance)\;
 \For{act in pred\_acts}{
  \eIf{the act doesn't need a slot}{
  add the single act into the list;
  }{
  pred\_slots = slot\_classifier(utterance, act)\;
  \For{slot in pred\_slots}{
    \eIf{the slot doesn't need a value}{
    add the act-slot pair into the list\;
    }
    {
    value = value\_decoder(utterance, act, slot)\;
    add the act-slot-value triple to the list\;
    }
    }
   }
 }
 Return the list containing triples.
 \caption{Hierarchical Decoding Algorithm}
 \label{alg:decoding}\textsf{}
\end{algorithm}

\subsection{Shared Utterance Encoder}
A bidirectional LSTM (BLSTM) \cite{schuster1997bidirectional, kawakami2008supervised} model is exploited to encode the utterance. Let $\bm{e}_w$ denote the word embedding of each word $w$, and $\oplus$ denote the vector concatenation operation. The encoder reads the utterance $\bm{w} = (w_1, w_2, \cdots, w_T)$ and generates $T$ hidden states of BLSTM: 
{
\begin{equation*}
\begin{split}
\bm{h}_i = \overleftarrow{\bm{h}_i} \oplus \overrightarrow{\bm{h}_i}; \quad \overleftarrow{\bm{h}_i} = f_l(\overleftarrow{\bm{h}_{i+1}}, \bm{e}_{w_i}); \quad \overrightarrow{\bm{h}_i} = f_r(\overrightarrow{\bm{h}_{i-1}}, \bm{e}_{w_i})
\end{split}
\end{equation*}
}where $\overleftarrow{\bm{h}_i}$ is the hidden vector of the backward pass in BLSTM and $\overrightarrow{\bm{h}_i}$ is the hidden vector of the forward pass in BLSTM at time $i$, $f_l$ and $f_r$ are LSTM units \cite{hochreiter1997long} of the backward and forward pass respectively. The final representation of the utterance (\emph{utterance vector}) is defined as:
{
\setlength\belowdisplayskip{1pt}
\begin{equation*}
\begin{split}
\bm{\Tilde{h}} = \overleftarrow{\bm{h}_1} \oplus \overrightarrow{\bm{h}_T}
\end{split}
\end{equation*}
}

The utterance vector $\bm{\Tilde{h}}$ will be used for the following act and slot type classifications, and hidden vectors $\{\bm{h}_1, \cdots, \bm{h}_T\}$ will be used for the value sequence generation with pointer network \cite{vinyals2015pointer, gulcehre-EtAl:2016:P16-1, see2017get}. 

\vspace{-2.5mm}
\subsection{Act and Slot Type Classifiers}
Act type prediction is defined as a multi-label classification problem here. A normal solution is to train a binary classifier for each label. We apply a feed forward network with two layers to calculate an existence score for each possible label:
{
\begin{equation*}
\begin{split}
\bm{r} & = \text{ReLU}(\bm{W}_u\bm{u}+\bm{b}_u) \\
\bm{p} & = \sigma(\bm{W}_r\bm{r}+\bm{b}_r)
\end{split}
\end{equation*}
}where $\bm{u}$ is the input vector, $\bm{W}_u, \bm{W}_r$ are weight matrices and $\bm{b}_u, \bm{b}_r$ are biases. $\sigma$ is the sigmoid function to normalize output scores. In the training stage, Binary Cross Entropy (BCE) loss function\footnote{Binary cross entropy loss suits multi-labels classification very well: $L=-\sum_i[y_i*log(p_i) + (1-y_i)*log(1-p_i)]$, where $y_i$ is the target value (0 or 1) of $i$-th label and $p_i$ is the predicted probability of $i$-th label.} is used. In the testing stage, classes with score higher than $0.5$ are predicted. For act type prediction, the input vector $\bm{u}$ is just the utterance vector $\bm{\Tilde{h}}$.

Slot type prediction is formatted in a similar way, while not only the utterance vector but also the corresponding act type are fed to the slot classifier. An embedding layer is also defined to encode each act type into a continuous vector. Let $a$ denote an act type and $\bm{e}_a$ denotes its embedding, then the input vector for the slot type classifier is:
{
\setlength\belowdisplayskip{1pt}
\begin{equation*}
\begin{split}
\bm{u}  = \bm{\bm{\Tilde{h}}} \oplus \bm{e}_a
\end{split}
\end{equation*}
}

A notable point is that we define embedding modules for act and slot types as word embedding to encode each type into a continuous representation. This allows us to utilize the predicted results from former modules in the latter, e.g., usage of act types in slot types prediction.

\subsection{Value Decoder with Pointer Network}

To predict value of the corresponding act-slot type pair, we utilize a sequence-to-sequence  model with attention \cite{luong2015effective} and pointer network \cite{vinyals2015pointer} to generate word sequence of the value. Since the encoder has been introduced above, we describe the details of the decoder below.

A LSTM model is used to decode the value sequence $\bm{v} = (v_1, v_2, \cdots, v_N)$. We define $v_N$ as \textit{``$<$/s$>$''} which means the end of a sequence. The LSTM proceed as $\bm{s}_i = f(\bm{s}_{i-1}, \bm{e}_{v_i})$, where $\bm{s}_i$ is the hidden vector at time $i$ and $f$ is the LSTM units. In order to incorporate the context information of corresponding act and slot, we define that:
{
\begin{equation*}
\begin{split}
\bm{\Tilde{s}}_i = \bm{W_s}(\bm{s}_i \oplus \bm{e}_a \oplus \bm{e}_s) + \bm{b_s}
\end{split}
\end{equation*}
}where $\bm{e}_a$ and $\bm{e}_s$ are embeddings of corresponding act type $a$ and slot type $s$ respectively, $\bm{W}_s$ is a weight matrix and $\bm{b}_s$ is a bias vector. $\bm{\Tilde{s}}_i$ is used in the attention mechanism to calculate context vector $\bm{c}_i$ as follows:
{
\begin{equation*}
\begin{split}
\bm{c}_i = \sum_{j=1}^T \alpha_{ij} \bm{h}_j; \quad \alpha_{ij} = \frac{\exp{( \bm{h}_j^T \bm{\Tilde{s}}_i )}}{ \sum_{k=1}^T \exp{( \bm{h}_k^T \bm{\Tilde{s}}_i )} }
\end{split}
\end{equation*}
} The encoded information of predicted act and slot in $\bm{\Tilde{s}}_i$ could help the attention mechanism to focus semantically.

Finally $\bm{\Tilde{s}}_i$ and $\bm{c}_i$ are concatenated to be an input of the output layer which calculates the probability distribution $\bm{P}_i^{gen}$ over the basic vocabulary as in \cite{luong2015effective}. 

To handle the OOV problem in value generation, we enhance the basic Seq2Seq model with pointer network \cite{vinyals2015pointer} which can generate a probability distribution $\bm{P}_i^{ptr}$ over the words of the input utterance according to the attention weights $\alpha_{ij}$. Therefore, the final distribution over the extended vocabulary (the basic vocabulary and words in the input utterance) is calculated as:
{
\setlength\abovedisplayskip{2pt}
\setlength\belowdisplayskip{1pt}
\begin{equation*}
\begin{split}
\bm{P}_i &= p_i * \bm{P}_i^{gen} + (1 - p_i) * \bm{P}_i^{ptr} \\
p_i & = \sigma(\bm{w_p}(\bm{e}_{v_i} \oplus \bm{\Tilde{s}}_i \oplus \bm{c}_i) + b_p)
\end{split}
\end{equation*}
}where $p_i$ is a balance score,  $\bm{w}_p$ is a weight vector and $b_p$ is a scalar bias.


\vspace{-4.5mm}
\section{EXPERIMENTS}

In our experiments, we use the dataset provided for the second Dialog State Tracking Challenge (DSTC2) \cite{henderson2014second}. It encompasses $11677, 3934, 9890$ pairs of utterance and the corresponding set of act-slot-value triples for training, development and testing respectively. Each utterance is annotated with semantics including multiple act-slot-value triples. Both the manual transcription and 10-best hypotheses are provided for each utterance. We use the manual transcription and top hypothesis (1-best) as inputs throughout our experiments. 

The dimension of embeddings is 100 and the number of hidden units is 128. Dropout rate is 0.5 and batch size is 20. Maximum norm for gradient clipping is set to 5 and Adam optimizer is used with an initial learning rate of 0.001. All training consists of 50 epochs with early stopping on the development set. We report F1-score of act-slot-value triples by the offical scoring script from http://camdial.org/~mh521/dstc/.

\textit{Glove}\footnote{\url{http://nlp.stanford.edu/data/glove.6B.zip}} word vectors are used to initialize all the embedding modules. For act and slot type embedding modules, we compose the embedding of these abstract concept words, for example, the embedding of \textit{``pricerange''} is the average of the embeddings of \textit{``price''} and \textit{``range''}. We also tie the act embedding and the topmost weight matrix of the act type classifier \cite{press2016using}, same for the slot embedding.

\vspace{-3.5mm}
\subsection{Overall Performance}
First of all, we conduct experiments on the top hypothesis and compare the results with prior arts to evaluate the overall SLU performance of our model. \cite{barahona2016exploiting, zhao2018improving} are neural network based methods which have been mentioned before, and \cite{williams2014web} is a statistical method which uses decision trees based binary classifiers to predict the presence of each slot-value pair and dialog act. The results are shown in Table \ref{table:state}. From the table, we can see that our model achieves the best F1 score and outperforms the prior works significantly.
\begin{table}[h]
\centering
\begin{tabular}{ |c||c| } 
 \hline
 Model & F1-score(\%)  \\ 
 \hline \hline
 SLU2 \cite{williams2014web} & 82.1 \\ 
  CNN+LSTM\_w4 \cite{barahona2016exploiting} & 83.6 \\ 
  S2S-Attn-Ptr-Net \cite{zhao2018improving} & 85.8 \\ 
 \hline 
 Our model & \textbf{86.9} \\
 \hline 
\end{tabular}
\caption{Comparison with published results on DSTC2.}
\label{table:state}
\end{table}

\vspace{-2.5mm}
\subsection{Generalization Capacity}
In this section, we would like to evaluate the generalization capacity of our hierarchical decoding (HD) model. We adopt two baselines for comparison. One baseline \cite{henderson2012discriminative} treats act-slot-value triple as a single label and train binary SVM classifiers to predict the existence of each label. The method is named as semantic tuple (ST) classifications. We replace the SVM with our BLSTM encoder for consistency. The other baseline \cite{yazdani2015model} proposes a model to calculate the similarity between the embeddings of input utterance and all possible semantic items (act-slot-value triples) by zero-shot (ZS) learning. We also apply a BLSTM encoder to get the utterance vector. We randomly select $5\%, 10\%, 20\%$ of the training data to create specific datasets with less act-slot-value triples. The testing set is the same as before. Both manual transcriptions and top hypotheses (1best) are used in the experiments. The results are shown in Table \ref{table:slu}. 

\begin{table}[h]
\centering
\begin{tabular}{ |c |c ||c |c |c| } 
 \hline
  Data Type & Data Size & ST & ZS   & HD \\ 
 \hline \hline
  \multirow{4}*{manual}    &   5\%   & 70.0 & 80.4 & \textbf{91.7}       \\ 
      & 10\%      & 92.4 & 91.0 & \textbf{93.6}      \\ 
      & 20\%      & 95.1 & 94.2 & \textbf{96.1}      \\ 
      & 100\%     & \textbf{98.3} & 98.1 & \textbf{98.3}      \\ 
  \hline
  \multirow{4}*{1best}    &   5\%     & 50.6 & 73.1   & \textbf{80.1}       \\ 
      & 10\%      & 80.0 & 79.6 & \textbf{80.5}      \\ 
      & 20\%      & 83.2 & 81.9 & \textbf{83.3}      \\ 
      & 100\%     & \textbf{87.2} & 86.1 & 86.9      \\ 
 \hline
\end{tabular}
\caption{SLU performance (F1) with varying training size.}
\label{table:slu}
\end{table}

As we can see, the performance of our model will not degrade heavily as the data becomes less and less. Especially, it achieves a much better F1 score than the baselines when only $5\%$ data remains. The results show that our model has a good capacity of generalization. 


To better explore the reason why our model achieves much better performance than the baselines with only $5\%$ data, 
we split the labels in the testing set into two categories according to whether the act-slot-value triple is seen in training set. Subsequently we report the F1 scores of our model and the baselines on these two categories in Table \ref{table:ratio}. 
We find that the decomposition of act-slot-value triples and hierarchical decoding of our method can generalize to unseen labels and improve the performance on seen labels simultaneously.
\begin{table}[h]
\centering
\begin{tabular}{ |c |c ||c |c |c| } 
 \hline
  Data Type & Label Type & ST  & ZS   & HD \\ 
 \hline \hline
\multirow{2}*{manual}    &   seen   & 71.1  & 81.2 & \textbf{92.2}       \\ 
      & unseen      & 0.0 & 10.7 & \textbf{72.6}      \\ 
  \hline
  \multirow{2}*{1best}    &   seen     & 51.1 & 74.2   & \textbf{80.8}       \\ 
        & unseen      & 0.0 & 5.2 & \textbf{45.3}      \\ 
 \hline
\end{tabular}
\caption{SLU performance (F1) with $5\%$ training data on two categories of labels (act-slot-value triples).}
\label{table:ratio}
\end{table}

\vspace{-5.5mm}
\subsection{Analysis}
The decomposition of act-slot structure allows us to predict unseen act-slot type pairs. For example, our model can predict \textit{``confirm(area)''} even if the pair does not exist in training set. Since it can learn to compose the semantics of \textit{``confirm(area)''} from   \textit{``confirm(food)''} and \textit{``inform(area)''}.

For non-enumerable slot types like \textit{``food''} and \textit{``name''} which may have a huge set of possible values, we can not define all the possible values in advance. The utilization of pointer network allows us to generate OOV values. In our experiments, most OOV values can be generated by recognizing the similar context around the values with pointer network. 

\begin{figure}[!htpb]
\centering
\includegraphics[width=0.75\linewidth,trim=.05cm .02cm .02cm .05cm, clip]{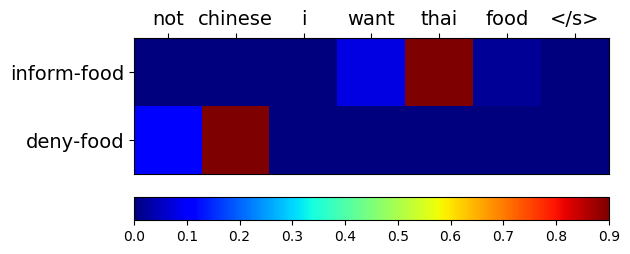}
\caption{Attention weights on input utterance of the value decoder with different act-slot pairs.}
\label{fig:attn}\textsf{}
\end{figure}

Given the predicted act and slot, the attention mechanism of the value decoder would focus on corresponding words. This enables the decoder to generate values accurately. Figure \ref{fig:attn} shows an example that how attention weights are distributed on the input utterance given different act-slot pairs. We can see that, ``\emph{inform-slot}" focuses on ``thai" and ``\emph{deny-slot}" concentrates on ``chinese" extremely.


\vspace{-5.5mm}
\section{CONCLUSION}

In this paper, we propose a novel hierarchical decoding model for SLU from unaligned data. The model exploits the structure of act-slot-value triples and can completely predict multiple triples. The decomposition of act-slot structure makes it possible to predict unseen act-slot type pairs. The utilization of pointer network in value decoder allows us to generate out-of-vocabulary (OOV) slot values. Finally, the experiment results show that our model possesses impressive performance and generalization capacity. In future, we would like to improve embeddings of act and slot types, and apply our method in domain adaptation problem of SLU.

\vfill\pagebreak

\bibliographystyle{IEEEbib}
\bibliography{refs.bbl}
\end{document}